\documentclass[10pt,twocolumn,letterpaper]{article}

\usepackage{authblk}
\usepackage[pagenumbers]{cvpr} 
\usepackage{graphicx}
\usepackage{amsmath}
\usepackage{amssymb}
\usepackage{booktabs}

\usepackage[pagebackref,breaklinks,colorlinks]{hyperref}

\usepackage[capitalize]{cleveref}
\crefname{section}{Sec.}{Secs.}
\Crefname{section}{Section}{Sections}
\Crefname{table}{Table}{Tables}
\crefname{table}{Tab.}{Tabs.}

\def\cvprPaperID{9947} 
\def\confName{CVPR}
\def\confYear{2022}

\begin{document}

\title{A Benchmark for Modeling Violation-of-Expectation in Physical Reasoning Across Event Categories}


\author[1]{Arijit Dasgupta}
\author[2]{Jiafei Duan}
\author[1]{Marcelo H. Ang Jr}
\author[3]{Yi Lin}
\author[4]{\\Su‑hua Wang}
\author[5]{Ren{\'e}e Baillargeon}
\author[2]{Cheston Tan}
\affil[1]{Department of Mechanical Engineering, National University of Singapore}
\affil[2]{Institute for Infocomm Research, A*STAR}
\affil[3]{Department of Psychology, New York University}
\affil[4]{Department of Psychology, University of California Santa Cruz}
\affil[5]{Department of Psychology, University of Illinois Urbana-Champaign}

\maketitle

\begin{abstract}

Recent work in computer vision and cognitive reasoning has given rise to an increasing adoption of the Violation-of-Expectation (VoE) paradigm in synthetic datasets. Inspired by infant psychology, researchers are now evaluating a model’s ability to label scenes as either expected or surprising with knowledge of only expected scenes. However, existing VoE-based 3D datasets in physical reasoning provide mainly vision data with little to no heuristics or inductive biases. Cognitive models of physical reasoning reveal infants create high-level abstract representations of objects and interactions. Capitalizing on this knowledge, we established a benchmark to study physical reasoning by curating a novel large-scale synthetic 3D VoE dataset armed with ground-truth heuristic labels of causally relevant features and rules. To validate our dataset in five event categories of physical reasoning, we benchmarked and analyzed human performance. We also proposed the Object File Physical Reasoning Network (OFPR-Net) which exploits the dataset's novel heuristics to outperform our baseline and ablation models. The OFPR-Net is also flexible in learning an alternate physical reality, showcasing its ability to learn universal causal relationships in physical reasoning to create systems with better interpretability.
\end{abstract}

\section{Introduction}
\label{sec:intro}
Physical-reasoning systems built on the foundations of intuitive physics and psychology are pivotal to creating machines that learn and think like humans \cite{lake2017building, adams2012mapping}. The ability to reason about physical events like humans opens a crucial gateway to multiple real-world applications from robotic assistants, autonomous vehicles, and safe AI tools. These systems are guided by facets of core knowledge \cite{spelke2007core}. Infant psychologists theorized that human newborns have an in-built physical-reasoning \cite{baillargeon2011infants, hespos2008young} and object representation \cite{kahneman1992reviewing,gordon1996s, stavans_lin_wu_baillargeon_2019} system. This has inspired researchers to approach the creation of physical-reasoning systems as a start-up software embedded with core knowledge principles \cite{ullman2020bayesian}.

Predicting the future of a physical interaction given a scene prior has been the most common task in designing computational physical-reasoning systems. One such task is the tower of falling objects, which has been as a test-bed for evaluating intuitive physics engines \cite{battaglia2013simulation, zhang2016comparative, lerer2016learning}. There have also been recent advancements in creating benchmarked datasets and models that combine multiple physical prediction tasks in a 3D environment \cite{duan2021space, bear2021physion, duan2021pip}. Physical reasoning has also been explored in Embodied AI \cite{duan2021survey} and Visual Question Answering (VQA) \cite{wu2017visual} with public datasets like \textbf{CLEVRER} \cite{yi2019clevrer}, \textbf{CRAFT} \cite{ates2020craft} and \textbf{TIWIQ} \cite{wagner2018answering} providing scenes of general and random interactions between objects and multiple questions concerning the physical outcome.

A parallel track complementary to future prediction is the design of artificial agents that can measure the plausibility of physical scenes. An agent capable of physical reasoning should not only be able to predict the future but also use it to recognize if a scene is \textit{possible} or \textit{impossible}. The Violation-of-Expectation (VoE) paradigm is an empirical diagnostic tool first implemented in infant psychology studies \cite{baillargeon1985object, baillargeon1987object} to measure the surprise of infants when shown \textit{possible} or \textit{impossible} scenes. The studies found that infants as young as 2.5 months could express surprise at a constructed scene that violated the principle of object permanence. This was akin to a magic show. VoE has since been used in an array of infant psychology experiments on a range of event categories in physical reasoning \cite{baillargeon1990young, spelke1995spatiotemporal, dan2000development, wang2005detecting, kotovsky1994calibration}.

The work in VoE has encouraged recent computational development of models and datasets \cite{piloto2018probing, riochet2018intphys, smith2019modeling, shu2021agent, gandhi2021bib} that challenge artificial agents to independently label \textit{possible} and \textit{impossible} scenes in physical events, goal preferences \cite{woodward1998infants} and more. While these datasets mimic real-world VoE experiments, they provide mainly vision data with little to no heuristics that aid learning. We believe that computational benchmarks in VoE require the embedding of more inductive biases from the body of psychology work that inspired them. A common finding among developmental psychologists on physical reasoning is that infants create abstract representations of objects \cite{kahneman1992reviewing,gordon1996s} from which they extract spatial and identity features. These high-level features are coupled with rules of reasoning that infants develop over time via process known as explanation-based learning \cite{baillargeon2017explanation} to form their expectation on how a physical scene should play out \cite{lin2020infants}.  Existing VoE datasets in physical reasoning lack such metadata in their scenarios. These metadata can be embedded into datasets to train models with greater interpretability and effectiveness in physical-reasoning tasks.

To this end, our contributions are three-fold. First, we present a new benchmark for physical reasoning, consisting of the first large-scale synthetic 3D VoE dataset with novel scene-wise ground-truth metadata of abstract features and rules. This dataset is inspired by findings in psychology literature and validated on human trials. Second, we propose a novel heuristic-based and oracle-based model framework to tackle the tasks of our VoE dataset. The model framework outperformed baseline and ablation computer-vision models. Third, we showed that our proposed model framework can learn an alternate reality of physical reasoning by leveraging on the feature and rule heuristics of the VoE dataset. This emphasises that the model is capable of learning universal causal relations in physical reasoning.

 \section{Related Works}


 
 The intersection of computer vision and physical reasoning is heavily grounded in the literature of VoE-based psychology. The age in which infants adopt core principles of persistence, inertia \& gravity \cite{lin2020infants} have been widely studied in a variety of event categories. For example, the \textbf{barrier} event is an event category illustrating (or violating) the constraint of solidity. In studies that implemented the barrier event to infants \cite{baillargeon1990young, spelke1992origins}, psychologists placed a solid barrier with an object on one side and the \textit{surprising} event occurred when infants were made to believe that the object could pass through the barrier. Like the barrier event, there are other events like \textbf{containment} where a \textit{surprising} scene was demonstrated with a taller object being fully contained in a shorter container, violating solidity constraints. Continuity constraints were also violated in \textbf{occluder} events \cite{baillargeon1991object, spelke1995spatiotemporal} when objects teleported behind one occluder to the back of another disconnected occluder. This event was also modified with different occluder heights in the middle segment \cite{aguiar19992}. Researchers also experimented with inertia violations in \textbf{collision} events \cite{kotovsky1994calibration, kotovsky1998development} where the outcome of the contact was impossible via linear momentum. Finally, gravity violations were also studied as \textbf{support} events \cite{baillargeon1992development, dan2000development, hespos2008young} where imbalanced objects that appeared stable were presented to infants.
 
 
 On the computational side of VoE-based physical reasoning, we find that Piloto et al. \cite{piloto2018probing}, IntPhys \cite{riochet2018intphys} and ADEPT \cite{smith2019modeling} to be the most relevant to our work, as they all employ the VoE paradigm in their datasets and evaluation. They all present 3D datasets of very similar event categories of physical reasoning.

\textbf{Piloto et al.} \cite{piloto2018probing} presents a 3D VoE dataset of 100,000 training videos and 10,000 pair probes of \textit{surprising} and \textit{expected} videos for evaluation. The dataset categorized their videos into `object persistence', `unchangeableness', `continuity', `solidity' and `containment'. Their Variational Autoencoder model benchmarked on the dataset and showed promise in `assimilating basic physical concepts'.

\textbf{IntPhys} \cite{riochet2018intphys} is a 3D VoE dataset with 15,000 videos of \textit{possible} events and 3,960 videos of \textit{possible} and \textit{impossible} events in the test and dev sets. Only three events on `object permanence', `shape constancy' and `continuity' were present. The study benchmarked the performances of a convolutional autoencoder and generative adversarial network with short and long-term predictions. The models performed poorly in comparison with their adult human trials but with higher than chance performance.

\textbf{ADEPT} \cite{smith2019modeling} is a model that
uses extended probabilistic simulation and particle filtering to predict object expectation. They use ADEPT on a 3D VoE dataset of 1,000 training videos of random objects colliding and 1,512 test videos of \textit{surprising} or \textit{control} stimuli. ADEPT accurately predicts the expected location of objects behind occluders to measure surprise, while replicating adult human judgements on the `how, when and what' traits of \textit{surprising} scenes \cite{smith2020fine}.

While these datasets provide vision data to replicate experiments in VoE for physical reasoning, they do not explicitly provide any heuristic-based metadata that models can exploit for higher-level interpretable predictions of physical reasoning. Researchers in cognitive AI have been calling for the use rule-based causal reasoning by adopting heuristics \cite{mottaghi2016happens, yi2019clevrer,chang2016compositional} in their approaches. We believe that inductive biases and heuristics that guide learning are important for computational physical reasoning. This is especially the case in VoE tasks that only train on \textit{expected} tasks and are made to predict which scenes are \textit{surprising} \cite{piloto2018probing,riochet2018intphys, smith2019modeling}. This motivated the construction of our VoE dataset.

\begin{figure*}
    \centering
    \includegraphics[width=\linewidth]{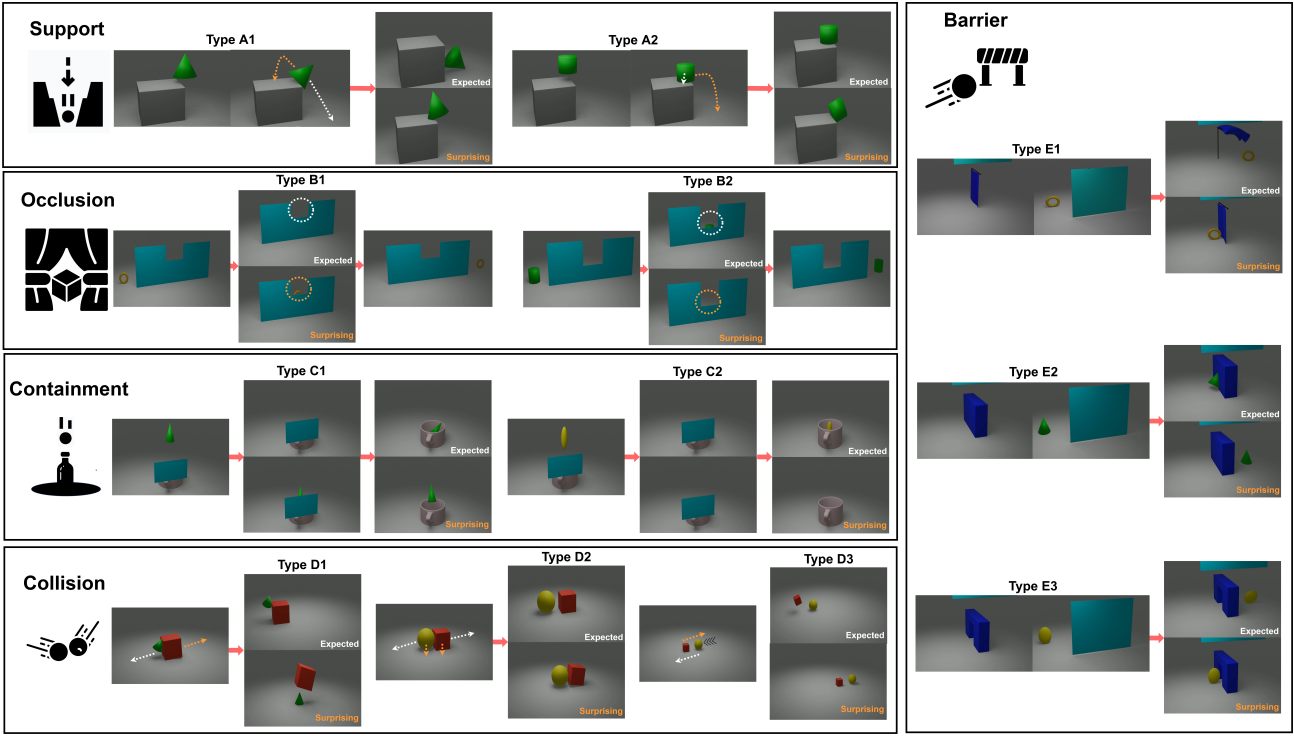}
    \caption{Examples of the different event categories in AVoE with their \textit{expected} and \textit{surprising} outcomes. \textbf{Support}: Type A1 (\emph{originally unbalanced}), Type A2 (\emph{originally balanced}). \textbf{Occlusion}: Type B1 (\emph{object shorter than occluder}), Type B2 (\emph{object taller than occluder}). \textbf{Containment}: Type C1 (\emph{Object fully contained in container}), Type C2 (\emph{object protruding out of container}). \textbf{Collision}: Type D1 (\emph{same speed, different size}), Type D2 (\emph{same speed, same size}), Type D3 (\emph{different speed, same size}). \textbf{Barrier}: Type E1 (\emph{soft barrier}), Type E2 (\emph{solid barrier}), Type E3 (\emph{barrier with opening}).}
    \label{fig:avoe_dataset}
\end{figure*}

\section{VoE Dataset}

\subsection{Overview}

Figure~\ref{fig:avoe_dataset} comprehensively illustrates the composition of the VoE dataset, which comprises synthetic video sub-datasets in five event categories: \textbf{support} (A), \textbf{occlusion} (B), \textbf{containment} (C), \textbf{collision} (D) and \textbf{barrier} (E). Each one of these event categories are split into further sub-categories that showcase physical variations based on the differing scene stimuli. They are described as follows.

\textbf{Support (Type A)}: An object is dropped above the edge of a support. The object's centre of mass is situated either over the edge (\textbf{Type A1}) or within the edge (\textbf{Type A2}).

\textbf{Occlusion (Type B)}: An inert object has initial momentum behind an occluder. The object can be shorter than the occluder's middle portion (\textbf{Type B1}) or taller (\textbf{Type B2}).

\textbf{Containment (Type C)}: An inert object falls from a short height above a container, with the interaction hidden behind an occluder. The object is short enough to be fully contained inside the container (\textbf{Type C1}) or tall enough to protude out of the container top (\textbf{Type C2}).

\textbf{Collision (Type D)}: Two inert objects with initial momentum collide head-on. In the first case (\textbf{Type D1}), two objects have the same initial speed with different sizes. In two other cases, both objects have similar size, with either the different (\textbf{Type D2}) or same (\textbf{Type D3}) initial speeds.

\textbf{Barrier (Type E)}: An inert object has an initial momentum to pass through a barrier, with their interaction hidden behind an occluder. To explore different barriers, the dataset comprises events with either a soft barrier (\textbf{Type E1}), a solid 
barrier (\textbf{Type E2}) or a barrier with opening (\textbf{Type E3}).

\subsection{Features and Rules}

Every event category $\psi \in \{A,B,C,D,E\}$ comes with sets of abstract features $f^\psi$, prior rules $r_{prior}^\psi$ \& posterior rules $r_{post}^\psi$ where $|f^\psi|=20$, $|r_{prior}^\psi|=13$ \& $|r_{post}^\psi|=9$. Prior rules are physical conditions about the event which can be answered with the features (e.g. height, width). These prior rules, coupled with the features, suffice to answer posterior rules that represent the outcome of the physical interaction. For instance, the features of a containment event could refer to the heights of the object and container. We present prior rules as a question: ``is the object taller than the container?'' to which the answer `yes' (based on a simple comparison of the features) would aid in answering a posterior rule as a question: ``did the object protrude out of the container?''. A full list of $f^\psi$, $r_{prior}^\psi$ \& $r_{post}^\psi$ is provided in the supplementary materials.

\subsection{Procedural Generation}
Multiple physical stimuli that affect the outcome of the interaction were randomly sampled to amplify the diversity of the dataset. Common parameters among the 5 sub-datasets included the object's shape $S_{Obj} \in \{Cube, Cylinder, Torus, Sphere, Cone, Side Cylinder, \\Inverted Cone\}$ and the object's height and width, $H_{Obj}, W_{Obj} \in [0.4, 1.6]$, where a $1$ is equivalent to 2$m$ in the 3D environment. The initial contact point of the object on the support in \textbf{A} $C_{Obj} \in [0.2, 0.8]$ where $C_{Obj} = 0.2$ indicates that the 20\% of the object's width is over the edge. In \textbf{B}, the occluder's middle segment height, $H_{Occ}, \in [0.1, 0.9]$ with 1 being the height of the occluder. In \textbf{C}, the container's shape, $S_{Con} \in \{Mug, Box\}$ was also varied with its height and width, $H_{Con}, W_{Con} \in [0.5, 1.5]$. The height ($\propto \text{mass}^{\frac{1}{3}}$) and initial speeds of objects in \textbf{D} were sampled as $H_{Obj} \in [0.5, 1.5]$ and $V_{Obj} \in [0.5, 2.5]$. In \textbf{E3}, the barrier's opening height and width were sampled in the range, $H_{Bar}, W_{Bar} \in [0.4, 1.4]$.

\subsection{Dataset Structure}

\begin{figure*}[]
    \centering
    \includegraphics[width=\linewidth]{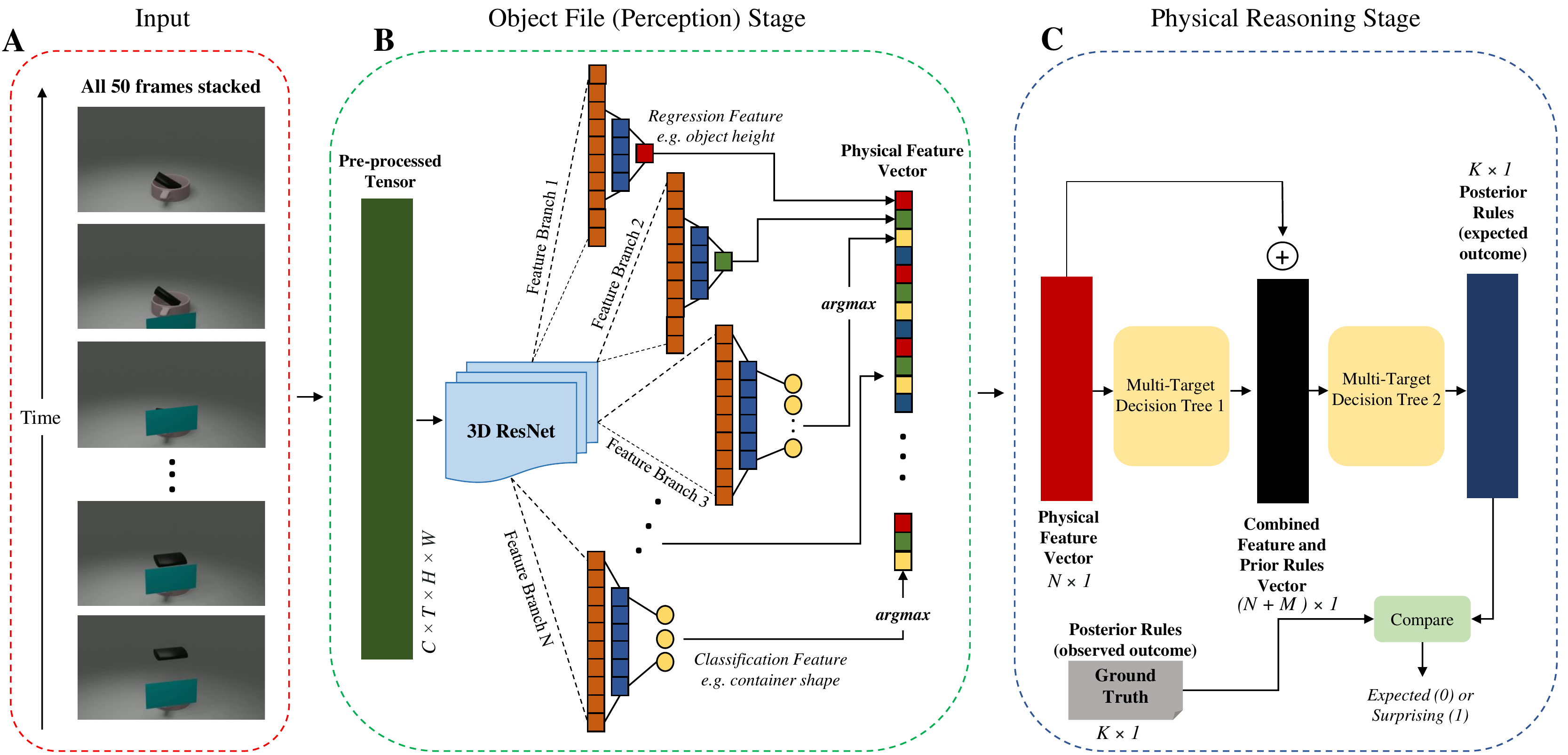}
    \caption{The OFPR-Net architecture. (A) \textbf{Input}: the original data inputs for the VoE task comprise 50 stacked frames. (B) \textbf{Object File (Perception) Stage}: The pre-processed input is fed into a 3D ResNet which then copies its output to $N$ feature branches. Each branch predicts either a scalar or classification feature. (C) \textbf{Physical Reasoning Stage}: The concatenated feature vector is fed into a multi-target decision tree to predict prior rules. The prior rules are concatenated with the feature vector and fed into another multi-target decision tree to predict the posterior rules. The predicted posterior rules are compared with the ground truth to determine if the input scene is \textit{surprising}.}
    \label{fig:ofpr_diagram}
\end{figure*}

Each event category $\psi$ has 5,000 different configured trials, amounting to 25,000 trials in the VoE dataset.  Every trial showcases an \textit{expected} or \textit{surprising} scene pair of the same stimuli. The training-validation-test dataset split is 75\%-15\%-10\%. This sums to 50,000 videos. At 50 frames per video, the VoE dataset offers 2,500,000 frames, each with a size of $960 \times 540$ pixels. The VoE dataset also provides the depth map and instance segmented frames. Along with the automatically generated ground-truth labels of $f^\psi$, $r_{prior}^\psi$ \& $r_{post}^\psi$ in every video, the frame-wise world position and orientation of all entities are provided. $f^\psi$, $r_{prior}^\psi$ \& $r_{post}^\psi$ are only used for training as they are not relevant to our VoE evaluation (see section~\ref{sec:metrics}). Nonetheless, they are still provided for the test and validation sets should researchers choose to evaluate performance in predicting $f^\psi$, $r_{prior}^\psi$ \& $r_{post}^\psi$. All frames were developed in the open-source 3D graphics software Blender \cite{blender}, using a Python API.

\section{OFPR-Net}

To understand how abstract features and rules can be used in physical reasoning, we examined a two-system cognitive model developed by infant psychologists \cite{stavans_lin_wu_baillargeon_2019}. They theorized that early stage physical reasoning is supported by an \textit{object-file} system \cite{kahneman1992reviewing} and a \textit{physical-reasoning} system \cite{baillargeon2011infants} that serve different functions. The \textit{object-file} system builds temporary spatio-temporal and identity representations of objects. When objects become involved in a causal interaction, the \textit{physical-reasoning} system becomes activated to predict the outcome of the interaction by first categorizing the event and then combining the temporary representations from the \textit{object-file} with its physical knowledge. If the observed outcome does not match the expected outcome, it is signaled as a \textit{surprising} event. To draw parallels between our VoE dataset and the \textit{object-file physical-reasoning} system, the features ($f^\psi$) are analogous to the temporary identity representations of objects recognized by the \textit{object-file} system. The prior and posterior rules ($r_{prior}^\psi$ \& $r_{post}^\psi$) are analogous to the symbolic structure of the \textit{physical-reasoning} system that provides its physical knowledge. We believe a simplified version of this two-system cognitive model can be computationally represented.

\textbf{To showcase how a model can exploit the novel heuristics on our dataset}, we introduce the Object File Physical Reasoning Network (OFPR-Net): a novel oracle-based model framework for modeling VoE in physical reasoning across event categories. A detailed architecture of the OFPR-Net is shown in Figure~\ref{fig:ofpr_diagram}. The essence of this model is to predict the expected outcome based on the stimuli of a scene, which can then be compared with the oracle of the actual outcome to decide if the scene is \textit{surprising}. The 50 RGB frames are stacked, pre-processed and propagated through two stages. First, the object file stage extracts all causally relevant features and their values. The data runs through a 3D ResNet and the output is copied into $N$ branches, representing $N$ causally relevant features. Each branch is either a feed-forward regression block that predicts a scalar value of a feature (e.g. object height), or a feed-forward classification block that predicts a categorical feature (e.g. object shape). As different event categories require a distinct set of features, the object file stage must also recognize which features are causally relevant. Therefore, all classification blocks are trained with an additional label for `irrelevant', signalling that the feature either does not exist (e.g. container height in an occlusion event, \textbf{Type B}), or it does not affect the outcome. All regression blocks are also trained to predict $-1$ in the event of an irrelevant feature. To avoid clashing with the $-1$ label, all scalar features are engineered to have positive values.

The $N$ concatenated features are propagated through the Physical Reasoning stage. First, the features run through a multi-target decision tree \cite{breiman2017classification} to predict the outcomes of $M$ prior rules. The feature vector is then concatenated with the predicted prior rule vector and is fed into a second multi-target decision tree to predict the $K$ posterior rules (expected outcome). Similar to the features, any irrelevant rules are classified as such. In this model, we assume that we have an oracle of the ground truth posterior rules, pointing to the actual observed outcome. The predicted posterior rules are compared with the ground truth. If they differ, the model signals the scene as \textit{surprising} and vice versa. As the focus is primarily on showing how the features and rules can be exploited to form expectations of the physical outcome and \textbf{not} on classifying the video's actual outcome, we found it suitable to use an oracle in this final step. This would pin the model's performance to the aforementioned focus.

\section{Experiments} 
\subsection{Human Trials}
\label{human_trials}

\begin{figure*}
    \centering
    \includegraphics[width=\linewidth]{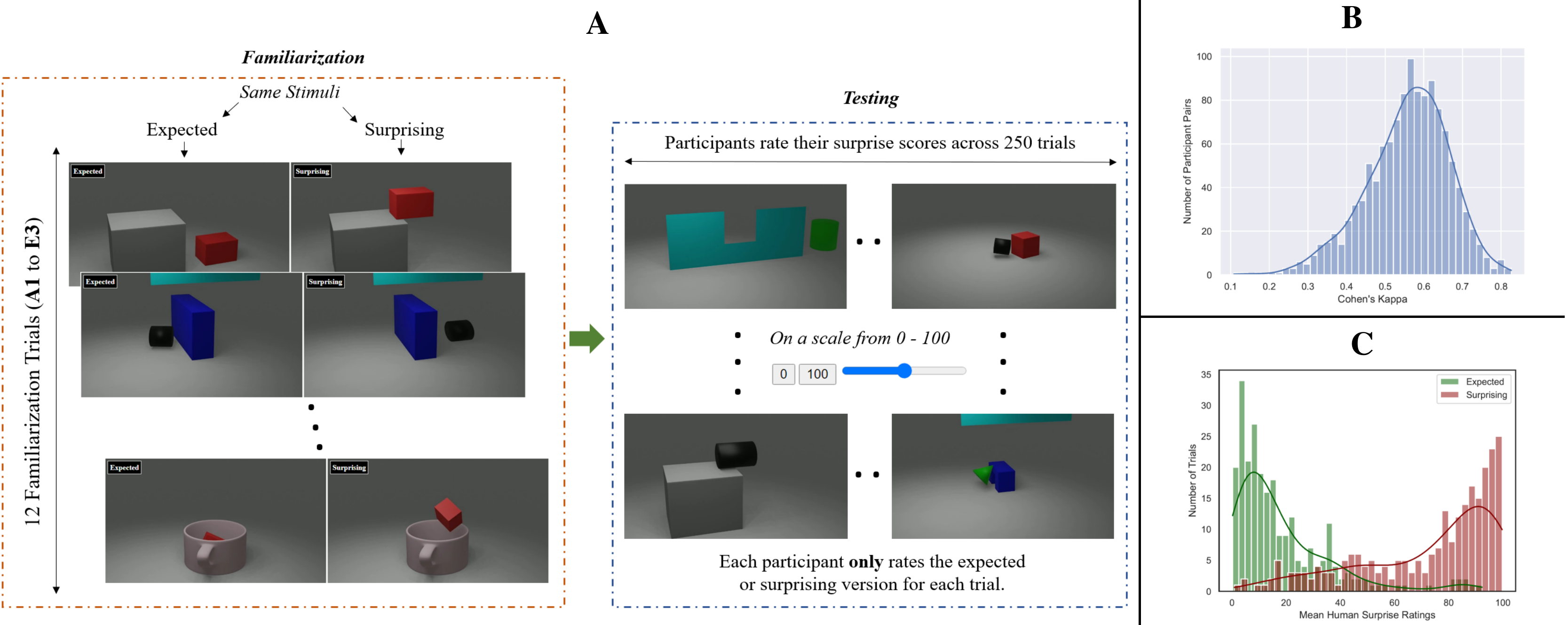}
    \caption{(A) Human trial setup with the familiarization stage and testing stage. (B) Cohen's $\kappa$ for common scene ratings (Z-scored) between all pairs of human participants. (C) Distributions of the mean human ratings per trial for \textit{expected} and \textit{surprising} scenes.}
    \label{fig:human_trials_diagram}
\end{figure*}

To benchmark human performance on the VoE dataset and validate its trials, we conducted an experiment testing adult humans on their judgement of the surprising level of the videos. 61 participants were recruited to answer an online questionnaire and were compensated with $\$7.50$ each.  Every participant was familiarized with 12 trials, where each trial showcased an \textit{expected} and \textit{surprising} version of the same stimuli. All participants were shown the same familiarization trials, and each trial represented a subcategory of every event category (\textbf{A1} - \textbf{E3}) and was selectively chosen from the training and validation sets. We randomly sampled 10\% of the combined VoE test set (250 trials $\leftrightarrow$ 500 scenes), drawn evenly from each event category. The task for each participant was to rate how surprising they found scenes assigned to them on an integer slider from 0 (\textit{expected}) to 100 (\textit{surprising}) as seen in Figure~\ref{fig:human_trials_diagram}(A). All responses were filtered via pre-set criteria for accuracy and consistency (see supplementary). 11 participants did not meet our pre-set criteria, hence we excluded their data. Therefore, we establish our analysis in the responses of 50 participants (31 female, 18 male and 1 non-binary) whose ages ranged from 19 to 32. Like the VoE human trials in \cite{smith2019modeling, shu2021agent}, the responses for each participant are standardized in the Z-normal distribution. This accounts for the participants' different usage of the slider, making their responses directly comparable.

Following the human trial methodology in  \cite{shu2021agent}, each participant was either shown the \textit{surprising} scene or the \textit{expected} scene of each trial. This is necessary to ensure each rating is independent and not influenced by comparing to another scene with the same stimuli. The number of \textit{expected} and \textit{surprising} scenes shown to each participant was evenly split. Hence, each participant was shown the \textit{expected} versions of 125 trials and the \textit{surprising} version of the other 125 trials. The \textit{surprising} and \textit{expected} versions of each trial were evenly split among all participants, such that half of all participants rated the \textit{surprising} version of each trial and vice versa. The human trials were conducted online via a custom web application made with Flask \cite{grinberg2018flask} to handle the back-end operations managing different scenes for each participant. 

\subsection{Baseline Model}

We establish a simple baseline by training a 3D ResNet \cite{hara3dcnns} pretrained on the Kinetics-700 dataset \cite{kinetics}. Four additional fully connected feed-forward layers are augmented to the 3D ResNet. These layers fine tune the model to output a scalar output with a sigmoid activation representing the surprise score from 0 (\textit{expected}) to 1 (\textit{surprising}). 
\subsection{OFPR-Net}

\textbf{Implementation:} The OFPR-Net is implemented using PyTorch \cite{paszke2019pytorch} and the \href{https://github.com/kenshohara/3D-ResNets-PyTorch}{3D ResNet} (MIT License) implementation by \cite{hara3dcnns} is used with their \href{https://drive.google.com/file/d/1fFN5J2He6eTqMPRl_M9gFtFfpUmhtQc9/view?usp=sharing}{pre-trained weights} (r3d34\_K\_200ep) on the Kinetics-700 dataset \cite{kinetics}. All regression blocks are trained with mean squared error loss and classifications blocks are trained with categorical cross-entropy loss. The model architecture assumes 24 features (engineered from 20 features to avoid negative scalar features), 13 prior rules and 9 posterior rules, matching what the VoE dataset offers. The multi-target decision trees are fully trained with $f^\psi$, $r_{prior}^\psi$ \& $r_{post}^\psi$ of the training set.

\textbf{Ablation study :} We conducted an ablation study to evaluate the inclusion of the Physical Reasoning stage of the OFPR-Net framework. Specifically, we removed the feature branches and the multi-target decision trees. The Object File stage is modified by replacing the output of the 3D ResNet with 9 classification block branches that attempt to predict the posterior rules (expected outcome). The model pipeline and implementation from the input to preprocessing and the 3D ResNet remain identical to OFPR-Net. Like the OFPR-Net, the predicted posterior rules are compared with the oracle of the ground truth posterior rules to determine if a scene is \textit{surprising} or \textit{expected}. As the Physical Reasoning stage is removed and the Object File stage remains as a modified version, we call this the OF-Net.

\subsection{Evaluation Metric}
\label{sec:metrics}
To evaluate model performance on the VoE dataset, we define the Hit Rate, $H_r$ ($= \sum^J_{i=1} H_r^i$), similar to \cite{riochet2018intphys} where $E^i$ and $S^i$ refer to the surprise level scores of the \textit{expected} and \textit{surprising} versions of trial $i$ from a set of $J$ trials. 

\begin{equation}
\label{hitrate}
      H_{r}^{i} =
    \begin{cases}
      1 & E^i < S^i\\
      0.5 & E^i = S^i\\
      0 & \text{Otherwise}
    \end{cases} 
\end{equation}

Unlike human responses, all models provide independent surprise ratings. Therefore, we feed both \textit{expected} and \textit{surprising} scenes with the same stimuli on the same model to compute $H_r$. The formulation of $H_r$ for human responses is adjusted to account for the fact that participants either rate the \textit{expected} or \textit{surprising} version of a trial, to ensure independent ratings. We consider all 625 ($25^2$) combinations of \textit{expected} scene ratings and \textit{surprising} scene ratings for each trial. The $H_r$ is computed by taking the average $H_r^i$ ($i$ now referring to the $i\text{th}$ combination) of the 625 combinations and then taking the mean of these average scores across all trials.

\begin{table*}[t]
\begin{center}
    \begin{tabular}{l|ccccc|ccc}
    \hline
     & \multicolumn{4}{c}{Hit Rate ($H_r$ for normal reality)}\\
    \hline
    Methods & {Support (\textbf{A})} & {Occlusion (\textbf{B})} & {Containment (\textbf{C})}  & {Collision (\textbf{D})} & {Barrier (\textbf{E})} & {Average}\\
    \hline\hline
    Human  & 0.686 & 0.844 & 0.946 & 0.788 & 0.883 & 0.829\\
    \hline\hline
    Random & 0.502 & 0.502 & 0.502 & 0.502 & 0.502 & 0.502\\
    Baseline& 0.500 & 0.500 & 0.500 & 0.500 & 0.500 & 0.500\\
    OF-Net (Ablation) & 0.629 & 0.818 & 0.811 & 0.491 & 0.745 & 0.699\\
    OFPR-Net (Ours) & \textbf{0.676} &  \textbf{0.907}& \textbf{0.855} & \textbf{0.532} & \textbf{0.768} & \textbf{0.748}\\
     
    \hline
    \end{tabular}
\end{center}
\caption{Hit Rate for Human Trials and all Models across all event categories. Best performing model is \textbf{bolded}.}
\label{table:final_results}
\end{table*}
\subsection{Experimental Setup}

\label{expt_setup}

The aim of the experiment in the present study is to compare the performance of models with humans in their ability to recognize if physical interactions within each event category is \textit{surprising} or expected. To fairly compare model performances with human performance, we evaluated all models with the exact 10\% test set used in the human trials (section \ref{human_trials}). To keep consistent with the testing size, all models were also trained and validated using 10\% of the training and validation sets. The experiments are split into 5 tasks: \textbf{A, B, C, D, E}. From \textbf{A} to \textbf{E}, every model is trained on data stipulated purely for each corresponding event category. Following the methodology of existing computational VoE datasets \cite{shu2021agent, smith2019modeling, piloto2018probing, riochet2018intphys, gandhi2021bib}, all models are \textbf{only trained on \textit{expected} videos}. This sums to 375 training scenes, 150 validation scenes and 100 testing scenes for each event category. Human trial responses for the same 100 test videos are used for comparison. Tackling these tasks for a model is challenging for the following two reasons: 1) The tasks are one-class classification problems. Learning to recognize anomalous \textit{surprising} scenes using only \textit{expected} scenes as training data requires a model to employ special heuristics and inductive biases to overcome the barrier of a purely imbalanced dataset. 2) Using 10\% of the full dataset forces the model to learn efficiently with limited data.

As the OFPR-Net, Ablation and Baseline models all have a 3D ResNet backbone, they undergo the same preprocessing steps. Each video frame is scaled to $112 \times 112$ and their individual elements scaled from 0 to 1. The input for each training sample maintained a shape of $3 \times 50 \times 112 \times 112$ (Color Channels, Frames, Height, Width) into the 3D ResNet. All pre-trained weights of the 3D ResNet backbone were frozen. Three major hyper-parameters (learning rate, batch size and optimizer choice) were optimized for each of the 6 task datasets via a grid search. A total of 20 grid search samples were extracted for each task and each sample was evaluated on the OFPR-Net for 30 epochs. The tuned hyper-parameters were used across all models for each task. Afterwards, each model is run on all tasks with 30 epochs for 10 seeded runs each. All models are run on a single NVIDIA Tesla V100-32GB GPU.

\section{Results and Analysis}
\subsection{Human Performance}

To check for inter-rater reliability, we measured the Cohen's $\kappa$ \cite{cohen1960coefficient} of our human responses. To classify each response, we assume that a Z score $<0$ indicates the \textit{expected} class and a Z score $\geq 0$ indicates the \textit{surprising} class. We believe that this is a fair assumption, as participants were shown an equal number of \textit{surprising} and \textit{expected} scenes and the Z-normal standardization accounts for the different usage of the rating slider. Given that the Cohen's $\kappa$ is measured between a pair of raters, we filtered out all common videos rated by each of the 1225 (=$50 \choose 2$) pairs of participants and measured the $\kappa$ based on them. Figure~\ref{fig:human_trials_diagram}(B) shows the uni-variate distribution of the $\kappa$ scores for all participant-pairs with a mean of 0.558. The distribution and the mean value reveal that the participants have `moderate' (close to `substantial`) agreement as defined by \cite{landis1977measurement}. 

Table~\ref{table:final_results} shows that the $H_r$ of the human responses are high in all event categories except for support (\textbf{Type A}). This is expected, as the support task was especially challenging for humans to tackle. As the view of the support was not perpendicular in the scene view, it is difficult to judge the position of the object mass center over the support edge precisely in the many corner cases where the object's mass center is very near to the support edge. The slight dip in performance for collision (\textbf{Type D}) can be explained in \cite{mitko2021striking} and \cite{kotovsky1994calibration}, showing that humans often mis-approximate the mass and the violation of object speed in collisions respectively.

The mean human rating spread based on the original rating values are shown in Figure~\ref{fig:human_trials_diagram}(C) to visualize the rating spread of \textit{expected} and \textit{surprising} scenes. The plot shows the participants can generally rate \textit{expected} and \textit{surprising} scenes accurately, regardless of the stimuli. This is further substantiated with an AUC-ROC \cite{fawcett2006introduction} score of 0.938 for the mean human ratings. The plot also illustrates only 6\% of \textit{expected} videos were rated $\geq 50$ on average, while 22.4\% of \textit{surprising} videos were rated $<50$ on average. This trend can be explained by considering the interpretation of a `surprise rating' to a human. Feedback gathered from the human trials revealed some participants would often set a low value for a scene they find surprising to distinguish from other scenes they find more surprising. This reinforces the idea that we cannot take 50 as a universal threshold and further justifies our decision to standardize the ratings.

\subsection{Model Performance}

Table~\ref{table:final_results} reveals the average $H_r$ performance of all models for 10 seeded runs for all event categories. The results show that OFPR-Net surpasses the performances of all models across all event categories. In particular, the OFPR-Net outperformed the OF-Net ablation model by an average of 7.01\%. This signals that the Physical Reasoning stage of the OFPR-Net boosts the performance in VoE tasks, showing the importance of learning features and their associations with the outcome via rules. However, the gap in performance is not very significant. This is not surprising, as the OF-Net still uses the posterior rule heuristics of our dataset to develop greater understanding of the expected outcome. For the sake of comparison, a random model that randomly selected a surprise rating in the uniform range $E^i,S^i \in [0,1]$ is shown in Table~\ref{table:final_results} to illustrate that the baseline is as good as random, performing significantly worse than models exploiting the heuristics in our dataset. By training on only \textit{expected} scenes, the baseline predicts all scenes as \textit{expected} (i.e. $E^i = S^i = 0$), hence receiving a consistent $H_r$ of 0.5. This baseline illustrates why a purely end-to-end model with no consideration of heuristics or inductive biases will not work on such tasks.

Comparing across the event categories, the OFPR-Net performed poorly in the collision (\textbf{Type D}) trials. Closer inspection of the $N$ feature branches losses of the Object File stage revealed that the model was poor in predicting the object velocities, which significantly altered the expected outcome in the Physical Reasoning stage of the OFPR-Net. As features like velocities require multiple frames to determine, it is more challenging to predict accurately with the limited data used for training. The human performance was higher than the OFPR-Net by an average of 10.83\% and performed better in all tasks except occlusion (\textbf{Type B}). Given the high standards of adult human physical reasoning, we find this acceptable. Therefore, surpassing human performance across all event categories remains an open challenge. Interestingly, our model's performance dipped in the same tasks where human performance dipped (\textbf{Types A \& D}). An explanation for this is that deeper level features like mass and centre-of-mass are difficult to infer just with vision as the main input modality, like with humans \cite{mitko2021striking}.

\subsection{Novel Insights}

The results confirm that the inherent structure of the OFPR-Net model can tackle the one class classification problem of the VoE dataset. By learning to predict features and their structures and relations with the rules, the OFPR-Net has added interpretability about the physical interaction. Not only can the OFPR-Net predict the expected outcome, but it can store knowledge the features and its basic causal relations to the outcome of the interaction. This added interpretability is crucial to creating safe AI applications. We believe that another advantage of the feature and rule based architecture of the OFPR-Net is that it is flexible to learn any reality presented to it. To test our hypothesis, we ran the OFPR-Net by \textbf{only training on the \textit{surprising} versions of all the training trials}. The model is made to believe that \textit{surprising} scenes depict reality and the \textit{expected} videos violate them. All other task-specific hyper-parameters and implementation were identical to the experimental setup (section~\ref{expt_setup}). In this flipped reality scenario, the hit rate is redefined as $1-H_r$ as the model should label \textit{expected} scenes more surprising than \textit{surprising} scenes.
\begin{figure}
    \centering
    \includegraphics[width=\linewidth]{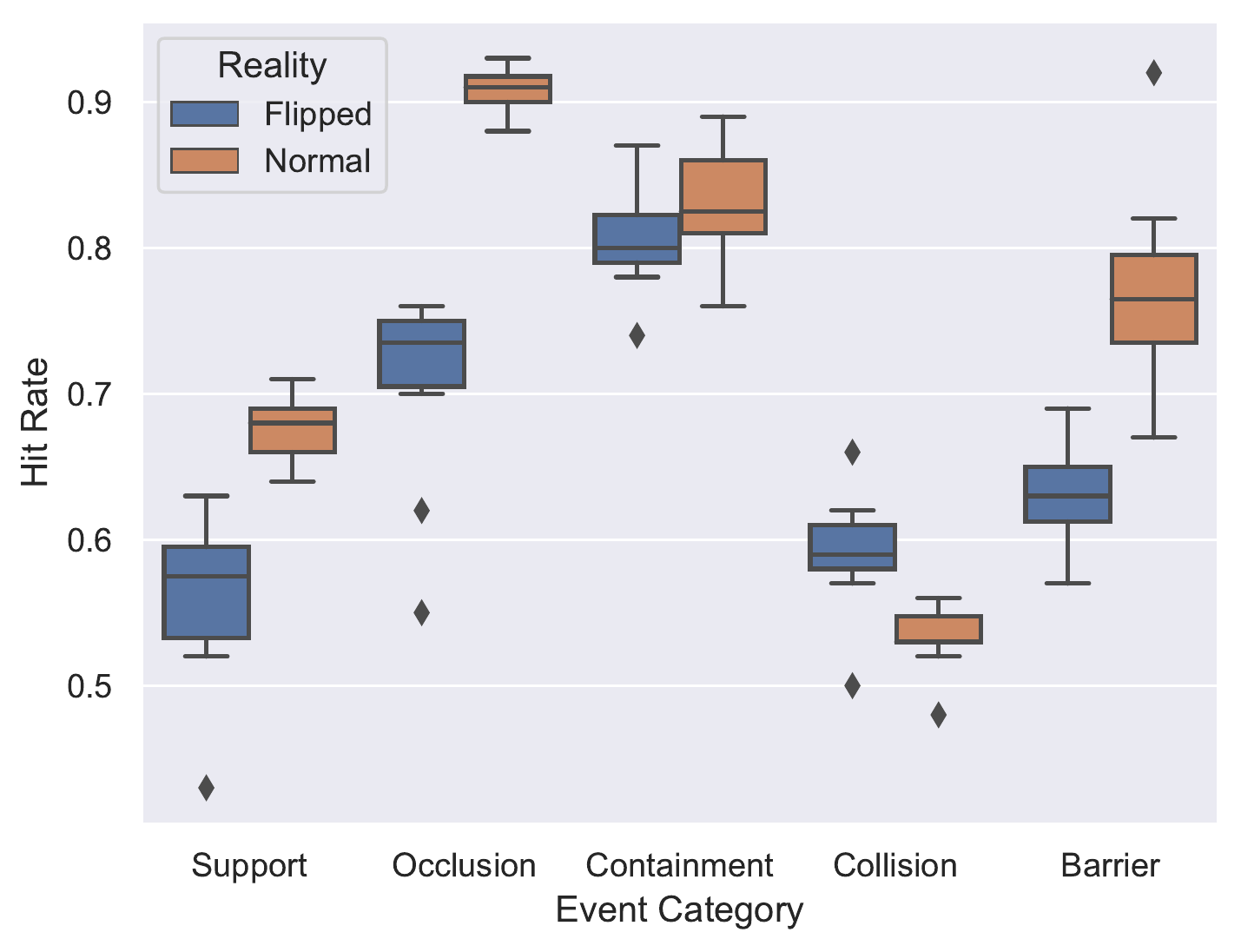}
    \caption{Box plots of Hit Rate for OFPR-Net when trained on only \textit{surprising} videos (flipped reality) or only \textit{expected} videos (normal reality) with the outliers shown. The performance was better than chance across all event categories in the flipped reality.}
    \label{fig:flipped_reality}
\end{figure}

While the OFPR-Net performs better in the normal reality (except collision, \textbf{Type D}), the box plots in Figure~\ref{fig:flipped_reality} show that it still performs reasonably and better than chance in the flipped reality. Hence, the OFPR-Net can treat \textit{surprising} videos as the new normal and is more likely to signal \textit{expected} videos as a violation in this new normal. The OFPR-Net signals that it is capable of learning universal causal relationships in physical reasoning. This is mainly possible because of the structural representation of $f^\psi$, $r_{prior}^\psi$ \& $r_{post}^\psi$ in the Physical Reasoning stage. To build systems that are capable of physical reasoning, frameworks that allow universal learning of causal relationships are crucial to ``support explanation and understanding, rather than merely solving pattern recognition problems" \cite{lake2017building}.

\section{Limitations and Future Work}

When implementing the human trials, each participant only rated either the \textit{expected} or \textit{surprising} version of a trial. While this takes precedence from previous work in computational VoE \cite{shu2021agent, smith2019modeling}, it meant that the human trial results are not directly comparable to the model output. Future computational VoE work with human trials may consider splitting the data collection into two stages with a set time period between the stages. The first stage can follow the method of our human trials, while the second stage shows the remaining videos not shown in the first stage. The time between the stages allows the participants to forget any stimuli, making the \textit{surprising} and \textit{expected} ratings independent.

One constraint of the dataset is that it only considers a few simple event categories and assumes each scene can only be represented by one event. In reality, physical interactions are much more complex, containing multiple event categories and a wider range of features and rules guiding the interactions. This scales up with more active objects in a rich environment. Nevertheless, we believe that providing $f^\psi$, $r_{prior}^\psi$ \& $r_{post}^\psi$ with scenes containing simple interactions is an important step to unveiling the potential of these heuristics. Future versions of the dataset will explore complex interactions with a wider range of rules and features. Frameworks built on our VoE dataset can also explore probabilistic and generative approaches to make use of the heuristics. These frameworks can learn to focus only on the relevant frames for efficient learning. An interesting approach would be to involve elements of unsupervised learning given that it is a one class classification problem. Finally, researchers may consider developing a curriculum-learning approach that attempts to closely replicate explanation-based learning \cite{baillargeon2017explanation} on VoE tasks. 

\section{Conclusion}

In this work, we showcase a novel approach to modeling VoE across basic event categories of physical reasoning. By leveraging on findings in the psychology literature, we proposed a novel synthetic 3D dataset augmented with ground-truth labels of abstract features and rules in five event categories of physical reasoning. The task of the dataset is to recognize scenes as \textit{expected} or \textit{surprising} with the added challenge of training on only \textit{expected} scenes. Human trials were conducted to benchmark human-level performance. The trials revealed that there was general agreement among participant responses. The participants were also proficient at rating \textit{surprising} videos with high surprise ratings and \textit{expected} videos with low surprise ratings. 

To showcase how using the abstract rules and features can tackle the challenge of our dataset, we proposed OFPR-Net, a novel oracle-based model framework inspired by a two-system cognitive model \cite{stavans_lin_wu_baillargeon_2019} of an infant's physical-reasoning system. The OFPR-Net benchmark surpassed the performance of the baseline and ablation models. However, average human-level performance still exceeded that of the OFPR-Net. Therefore, it remains an open challenge to beat the benchmarked human-level performance on our dataset. Finally, we show that the structural nature of the OFPR-Net guided by features and rules is flexible in learning an alternative reality of physical reasoning. This emphasises that a model that exploits heuristics of physical reasoning is capable of learning universal causal relations that are necessary to create systems with better interpretability. This validates the novelty of our dataset and encourages future work in this paradigm to focus on such heuristics and inductive biases for learning in physical reasoning. 

\section*{Acknowledgements}
This research is supported by the Agency for Science, Technology and Research (A*STAR), Singapore under its AME Programmatic Funding Scheme (Award \#A18A2b0046). We
would also like to thank the National University of Singapore for the computational resources used in this work.

{\small
\bibliographystyle{ieee_fullname}
\bibliography{egbib}
}

\end{document}